\documentclass[nonatbib]{article}

\usepackage[preprint]{neurips_2024}

\usepackage[utf8]{inputenc} 
\usepackage[T1]{fontenc}    
\usepackage{hyperref}       
\usepackage{url}            
\usepackage{booktabs}       
\usepackage{amsfonts}       
\usepackage{nicefrac}       
\usepackage{microtype}      
\usepackage{xcolor}         
\usepackage{colortbl}
\usepackage{amsmath}
\usepackage{enumitem}
\usepackage{graphicx}
\usepackage{multicol}
\usepackage{multirow}
\usepackage{pifont}
\usepackage{svg}
\usepackage{tabularx}
\usepackage{todonotes}
\usepackage{wrapfig}

\newcommand{\cellc}{\cellcolor{lightgray!50}}

\title{RTGen: Generating Region-Text Pairs for Open-Vocabulary Object Detection}

\author{Fangyi Chen\thanks{The first two authors contributed equally.} \quad Han Zhang$^*$ \quad Zhantao Yang \quad Hao Chen \quad Kai Hu \quad Marios Savvides\\
Carnegie Mellon University\\
{\tt\small \{fangyic,hanz3,zhantaoy,haoc3,kaihu,marioss\}@andrew.cmu.edu}
}

\begin{document}

\maketitle

\begin{abstract}
Open-vocabulary object detection (OVD) requires solid modeling of the region-semantic relationship, which could be learned from massive region-text pairs. However, such data is limited in practice due to significant annotation costs. In this work, we propose \textbf{RTGen}
to \underline{gen}erate scalable open-vocabulary \underline{r}egion-\underline{t}ext pairs and demonstrate its capability to boost the performance of open-vocabulary object detection. RTGen includes both text-to-region and region-to-text generation processes on scalable image-caption data. The text-to-region generation is powered by image inpainting, directed by our proposed scene-aware inpainting guider for overall layout harmony. For region-to-text generation, we perform multiple region-level image captioning with various prompts and select the best matching text according to CLIP similarity. To facilitate detection training on region-text pairs, we also introduce a localization-aware region-text contrastive loss that learns object proposals tailored with different localization qualities. Extensive experiments demonstrate that our RTGen can serve as a scalable, semantically rich, and effective source for open-vocabulary object detection and continue to improve the model performance when more data is utilized, delivering superior performance compared to the existing state-of-the-art methods. The code and data will be released at \url{https://github.com/seermer/RTGen}

\end{abstract}

\section{Introduction}
\label{sec:intro}

Deep learning models trained on sufficient defined-vocabulary data are effective in solving object detection tasks \cite{girshick2014rich, ren2015faster, redmon2016you, lin2017focal, tian2020fcos, carion2020end, li2023AAG, chen2023enhanced, li2023robust, Zhu2019SoftAO}, but in the open world, detecting thousands of object categories remains a challenge. Great interest has been raised in improving open-vocabulary object detection (OVD), which is expected to detect objects of arbitrary novel categories that have not necessarily been seen during training \cite{gu2021open, du2022learning, kuo2022f, zang2022open, ma2024codet, xu2023exploring, kamath2021mdetr, long2023capdet}. 

Recently, the advancements in vision-language models \cite{radford2021learning, jia2021scaling, cherti2023reproducible} have propelled image-level open-vocabulary tasks to new heights through the utilization of contrastive learning across a vast scale of image-caption pairs \cite{sharma2018conceptual, thomee2016yfcc100m, schuhmann2022laion}. This progress inspires researchers in the detection community to explore OVD methods drawing upon similar principles. However, training object detector needs region-level annotations. Unlike web-crawled image-caption pairs, region-level instance-text (region-text) pairs are limited and expensive to annotate, which leads us to two fundamental questions: \textit{how to acquire scalable region-text pairs}, and \textit{how to effectively utilize them for training OVD}?

\begin{figure*}[t]
    \centering
    \includegraphics[width=1.0\columnwidth]{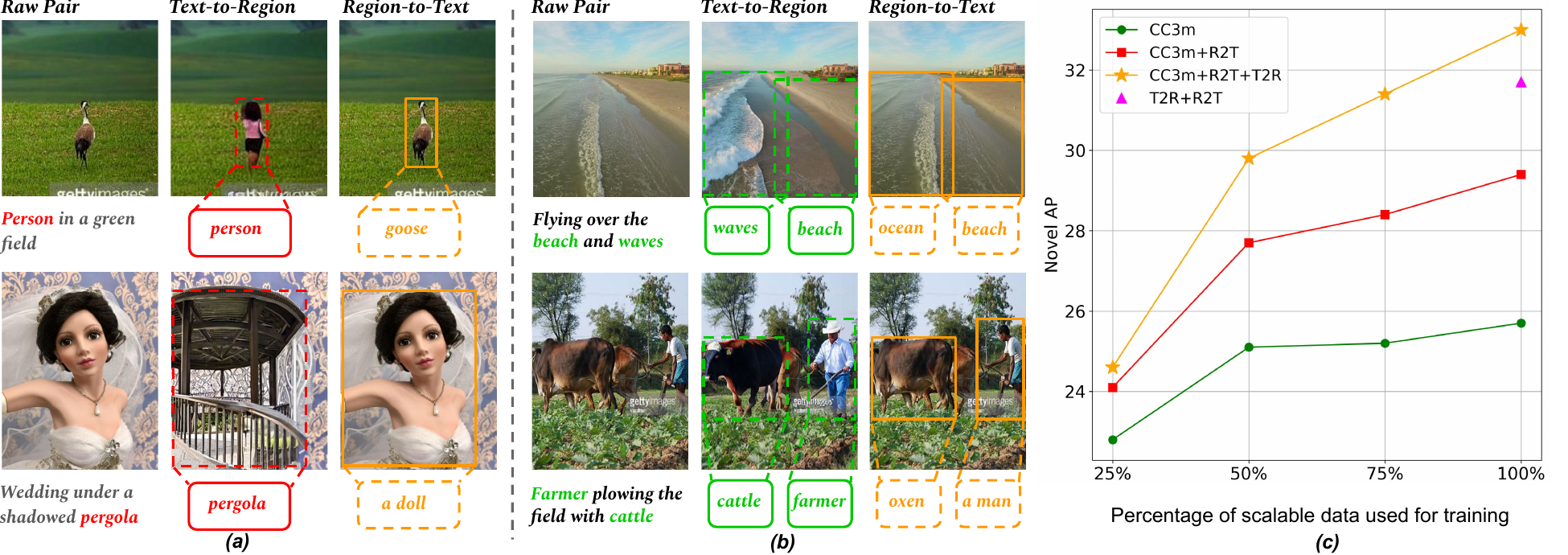}
    \caption{ 
    (a) Image-caption pairs may lack visual-textual correlation, addressed through the text-to-region (T2R) and region-to-text (R2T) processes.
    (b) The generative processes introduce a breadth of visual and textual diversity.
    (c) The average precision on OV-COCO novel classes when training with different percentages of our generated data which
    demonstrates a \textit{steep} increase when R2T+T2R is applied, evidencing a marked improvement proportional to the volume of data utilized
    .}
    \label{fig:fig1}
\end{figure*}

Some recent approaches focus on acquiring region-level pseudo labels by mining structures or data augmentation from image-caption pairs \cite{zhou2022detecting, ma2024codet, lin2022learning, li2022grounded, minderer2022simple, wu2024clim}. They are typically designed to align image regions with textual phrases extracted from corresponding captions. This is achieved by either leveraging a pre-trained OVD model to search for the best alignment between object proposals and phrases\cite{li2022grounded, liu2023grounding, zhao2022exploiting}, or through associating the image caption with the most significant object proposal \cite{zhou2022detecting, lin2022learning}. However, these web-crawled data may lack of accurate image-caption correspondence as many captions do not directly convey the visual contents, as shown in Fig.\ref{fig:fig1}(a). 
In addition, the precision of alignment is significantly dependent on the performance of the pre-trained OVD models, resulting in a recursive dilemma: a good OVD detector is requisite for generating accurate pseudo predictions, which in turn are essential for training a good OVD detector. 



Different from prior arts, this paper studies to leverage generative models to synthesize a rich corpus of region-text pairs for training OVD.
Unlike OVD models whose training relies on limited detection/grounding data, generative models \cite{rombach2022high, li2023gligen, zhang2023adding, li2022blip, li2023blip, Li2023generation} are typically trained on extensive datasets that have both imagery and textual modalities. 
Within the framework outlined in this study, generated open-vocabulary region-text pairs are proven as effective in accomplishing the task.

More specifically, our generation process is rooted in the web-crawled image-caption pairs and operates under two paradigms: text-to-region (T2R) and region-to-text (R2T). In the text-to-region process, we guide a diffusion model \cite{li2023gligen} to execute the inpainting,
conditioned on extracted caption phrases and image-predicted proposal boxes.
A key design of this process is the allocation of phrases and boxes to achieve overall layout harmony. This is facilitated by training a novel \textit{Scene-Aware Inpainting Guider (SAIG)}, designed to interpret the multi-modal scene comprehensively and sample flexible layouts that guide the inpainting within contextually relevant and geometrically coherent regions.
In the region-to-text process, we verify that simply applying a powerful captioning model \cite{li2023blip} on object proposals is an effective way to generate region-text pairs. 
The generation exhibits three novel characteristics: 
\textbf{Firstly}, rather than applying generative models on pre-existing detection datasets \cite{suri2023gen2det, feng2024instagen, Cho2023OpenVocabularyOD}, our generation is based on image-caption pairs that are scalable and mirror the real-world distribution, aligning well with the nature of open-vocabulary setting. \textbf{Secondly}, unlike some concurrent researches \cite{feng2024instagen, Cho2023OpenVocabularyOD} directly work on the novel categories, we structure the generation process under the circumstance of without knowing the novel categories in advance.  
\textbf{Thirdly}, models from two distinct domains introduce a breadth of semantic richness and knowledge, enhancing the diversity of the generated data, as shown in Fig.\ref{fig:fig1}(b) and (c). 
As the capabilities of generative models continue to advance, we anticipate further enhancements in this framework over time.

To effectively use the generated region-text pairs, 
we expand contrastive learning to fit detection learning scenarios, by incorporating not only the generated region-text pairs but also the adjacent, less accurate regions to learn with dynamic targets and weights. This loss function, termed Localization-Aware Region-Text Contrastive Loss, can be integrated into the training pipeline of various object detectors, allowing for joint training with standard detection data.
Our contributions can be summarized in three folds:
\begin{itemize}[leftmargin=2em]
\item We propose a framework that generates open-vocabulary region-text pairs from image-caption pairs. The framework features a text-to-region process, which is the first attempt to synthesize region-text pairs for training OVD without prior knowledge of the novel categories, as well as a region-to-text process that populates the generation with abundant regional captions.


\item We introduce a novel Scene-Aware Inpainting Guider to facilitate text-to-region generation. We 

also introduce a new loss function, as an expansion of contrastive loss, which enables detectors to effectively learn from generated region-text pairs. 

\item We apply our methods on Faster-RCNN \cite{ren2015faster} and CenterNet2 \cite{zhou2021probabilistic} and conduct extensive ablation studies and experiments on OV-COCO and OV-LVIS benchmarks. Without bells and whistles, training with RTGen brings +\textbf{6.9} box $\text{AP}_{novel}$ on OV-COCO and +\textbf{4.2} mask $\text{AP}_{novel}$ on OV-LVIS compared to our baseline, and outperforms other state-of-the-art methods.
\end{itemize}

\section{Related Work}
\subsection{Open Vocabulary Object Detection}
Open-vocabulary Object Detection (OVD) learns to generalize from annotated (seen) object categories to any (unseen) categories imagery. Different from Zero-shot Object Detection (ZSD), OVD is typically trained on an instance-level annotated datasets and meanwhile allowed the flexibility to learn information from open-world text corpora and images \cite{zareian2021open}. A recent trend tries to adapt the pretrained vision-language models (VLMs), such as CLIP\cite{radford2021learning}, to object detectors in attempt of zero-shot generalization capabilities. For example, ViLD\cite{gu2021open} directly distills the visual and textual knowledge of CLIP to the object detectors. However, because these VLMs are typically trained on image-text pair data, deploying them directly can be suboptimal for region-level recognition. As a result, many works have been searching to establish a solid region-text alignment. RegionCLIP\cite{zhong2022regionclip} proposes to first finetune CLIP to learn region-text alignments and then transfer CLIP to OVD. DetPro\cite{du2022learning} incorporates a trainable text prompt to align the VLM to the detection task. Similarly, PromptDet\cite{feng2022promptdet} aims to align the textual embedding space with regional visual object feature through region-level prompting. Alternatively, several works examine more sophisticated approaches to mine the structural correspondence. VLDet\cite{lin2022learning} proposes to utilize a bipartite set-matching algorithm to find region-text pairs through the image. CoDet\cite{ma2024codet} finds the common regions of interest among many images by constructing image-caption groups that may contain similar visual contents. In addition, hand-crafted region-text alignment by simple data augmentation are proven as effective. Detic\cite{zhou2022detecting} directly learns the alignment between the text labels and the max-sized object proposals. BARON\cite{Wu2023AligningBO} and CLIM\cite{wu2024clim} augment the region proposals by grouping and the mosaic arrangement are shown to be effective approaches. In this work, we tackle the problem from a perspective of generated region-text pairs.

\subsection{Training Object Detection with Synthetic Data}

For generated image data, prior works utilizes model-free data augmentation techniques such as simple copy-paste\cite{ghiasi2021simple}. For an open-vocabulary detection task, these methods are strongly limited by the lack of semantic diversity and instance-level variations. Gen2Det\cite{suri2023gen2det} proposes to use grounding-based image generation to acquire scene-centric images to enhance the detectors' ability on rare object categories in a long-tail categorical distribution. Similarly, MosaicFusion\cite{Xie2023MosaicFusionDM} borrows an off-the-shelf diffusion model to generate object-level images and mask annotations. InstaGen\cite{Feng2024InstaGenEO} aims to enable novel category generalization by generating images that contains both base and novel object categories as well as their bounding boxes. In general, these generation models assume knowing the novel object categories in advance. On the contrary, we propose to generate images under the same open-world setting to minimize the leaking of information. CapDet\cite{long2023capdet} trains a dense captioning head as an auxiliary learning objective. PCL\cite{Cho2023OpenVocabularyOD} proposes to train an image captioning model on the image-caption pairs and then generate region-level captions. However, they typically work under a finite dataset that are not easily scalable.  

\section{Generating Region-Text Pairs for Training OVD}
\subsection{Preliminaries}
\label{sec:prelim}
\textbf{The OVD problem and our goal:}
Open-vocabulary object detection is first proposed in OVR-CNN\cite{zareian2021open}. Formally, a detector is trained on a detection dataset with a predefined set of base object categories $\mathcal{C}\textsuperscript{base}$. 
Meanwhile, it leverages external image-caption pairs with an abundant list of vocabulary $\mathcal{C}\textsuperscript{open}$. 
During testing, the detector is expected to detect arbitrary novel object categories $\mathcal{C}\textsuperscript{novel}$, where $\mathcal{C}\textsuperscript{base} \cap \mathcal{C}\textsuperscript{novel} = \emptyset$. In a strict open-vocabulary setting, $\mathcal{C}$\textsuperscript{novel} are only known in testing.


Given an image-caption pair, our goal is to generate a set of region-text pairs 
$\{(r_j, t_j)\}_{j \in [N]}$ , 
where $r_j$ denotes a region on image bordered by a bounding box, and $t_j$ denotes the text (phrase) that semantically aligns with $r_j$. 
Subsequently, we expect to use the region-text pairs to train the open-vocabulary object detectors.

\begin{figure*}[t]
    \centering
    \includegraphics[width=1.0\columnwidth]{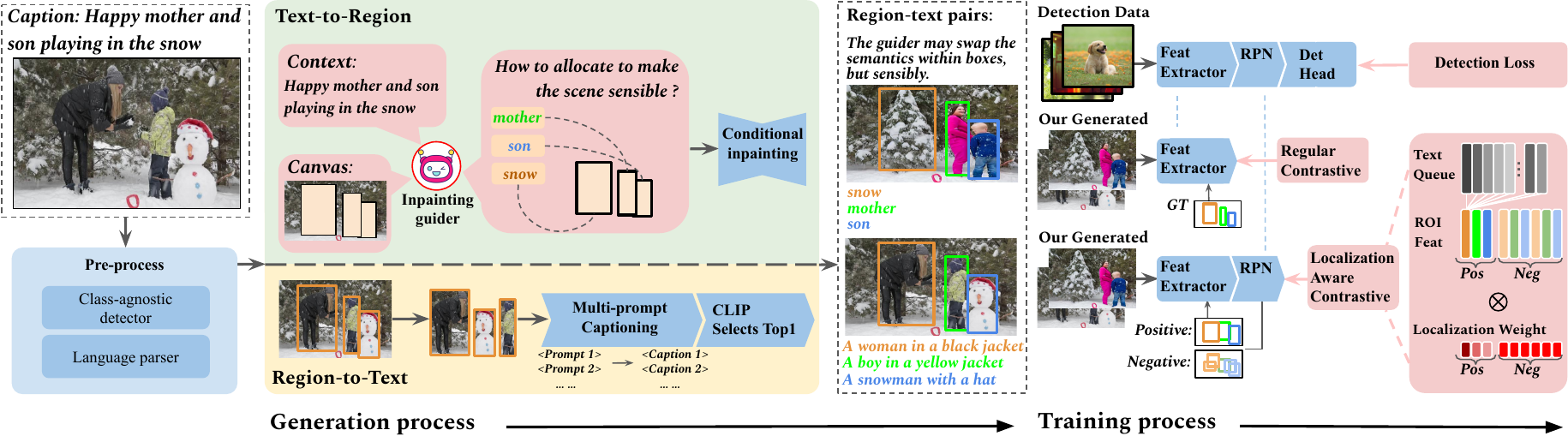}
    \caption{Framework overview. We generate region-text pairs from image-caption pairs through T2R and R2T processes. Detectors are jointed trained via localization-aware region-text contrastive loss. 
    }
    \label{fig:overview}
\end{figure*}

\textbf{Framework overview.} Illustrated in Fig.\ref{fig:overview}, we start from image-caption pairs with two basic pre-processings: a class-agnostic detector is applied to the image to produce proposal boxes; a large language model \cite{jiang2023mistral} is employed on the caption to identify tangible and physical phrases (see Appendix \ref{sec:Appendix}). They are subsequently input into the generation framework, where text-to-region is executed by an inpainting guider followed by an inpainting model (Section \ref{sec:t2r}), while region-to-text is conducted by a captioning model and a subsequent selection step (Section \ref{sec:r2t}). The generated region-text pairs incorporate contrastive learning and the proposed localization-aware region-text contrastive loss (Section \ref{sec:lart}), jointly training with detectors. 

\subsection{Text-to-Region}
\label{sec:t2r}
Image-caption pairs ensure the generation inherits visual and semantic richness. Although generating images from texts with the controllability of layout has been widely researched in recent years \cite{feng2024layoutgpt, qin2024diffusiongpt, li2023gligen, zhang2023adding}, the generation of image regions from image-caption pairs remains underexplored. 

\subsubsection{Inpainting Image-Caption Pairs.}
Naturally, we build upon image inpainting as it directly gives region-text alignment while preserving a substantial proportion of the original image, thus transferring the realism and diversity of the images to the generated output, particularly in the context regions, which is critical within the setting of open-vocabulary detection. Considering an image $I$, a phrase $t$, and a specified proposal box $b$. An inpainting model, denoted as $\mathcal{G_I}$, can replace the original visual content inside $b$ (region $r$) with a newly generated region $\hat{r}$, 

\vspace{-0.1in}
\begin{equation}
\label{eq:2}
    \hat{r} = \mathcal{G_I}(I, (b, t))
\end{equation}
\vspace{-0.1in}


\MakeLowercase{where} $\hat{r}$ is aligned with $t$ semantically, while the rest of the image $ I \setminus r$ remains unchanged. 

When inpainting an image-caption pair, 
we have acquired $N$ proposal boxes $\mathcal{B}=\{b_1, b_2, ..., b_N\}$ from the image and $M$ extracted phrases $\mathcal{T}=\{t_1, t_2, ..., t_M\}$ from the caption by the pre-processing. Here, a preliminary step is to allocate proposal boxes and phrases to get a harmonious layout. We recognize several characteristics of this task: (1) There exists $t$ that is not related to any region in the image, and vice versa.
(2) A box $b$ could be of any shape and located in any context, yet in natural images, regions with a semantic meaning may follow certain geometric distributions. 



With these considerations, we desire a new approach to \textbf{scene-aware allocation for inpainting} that can sample a harmonious layout by allocating the $\mathcal{B}$ and $\mathcal{T}$, based on its understanding of the scene.  As an example shown in Fig.\ref{fig:saig}, the core challenge is to understand "Happy mother and son playing in the snow" and allocate the phrases "mother", "son", and "snow" to the proper boxes. 
It is worth noting that, for obtaining region-text pairs, it's unnecessary to inpaint a region with a phrase that replicates its original visual content (e.g., by grounding). Instead, the preferred design is that the inpainting process will flexibly conform to a distribution that is contingent on the scene's context and is consistent with what is typically observed in the real world. We also tried random allocation and grounding-based allocation and analyzed their drawbacks in Section \ref{sec:exp}.

\subsubsection{Scene-Aware Inpainting Guider (SAIG)}

We model the probability of allocating a pair ($b_n$, $t_m$) as a joint probability $p_{nm} = P(b_n, t_m | \text{scene})$, which is decomposed equally:

\begin{equation}
\label{eq:4}
    P(b_n, t_m | \textrm{scene}) = P(t_m | b_n, \textrm{scene}) \times P(b_n | \textrm{scene}) 
\end{equation}




\begin{figure*}[t]
    \centering
    \includegraphics[width=1.0\columnwidth]{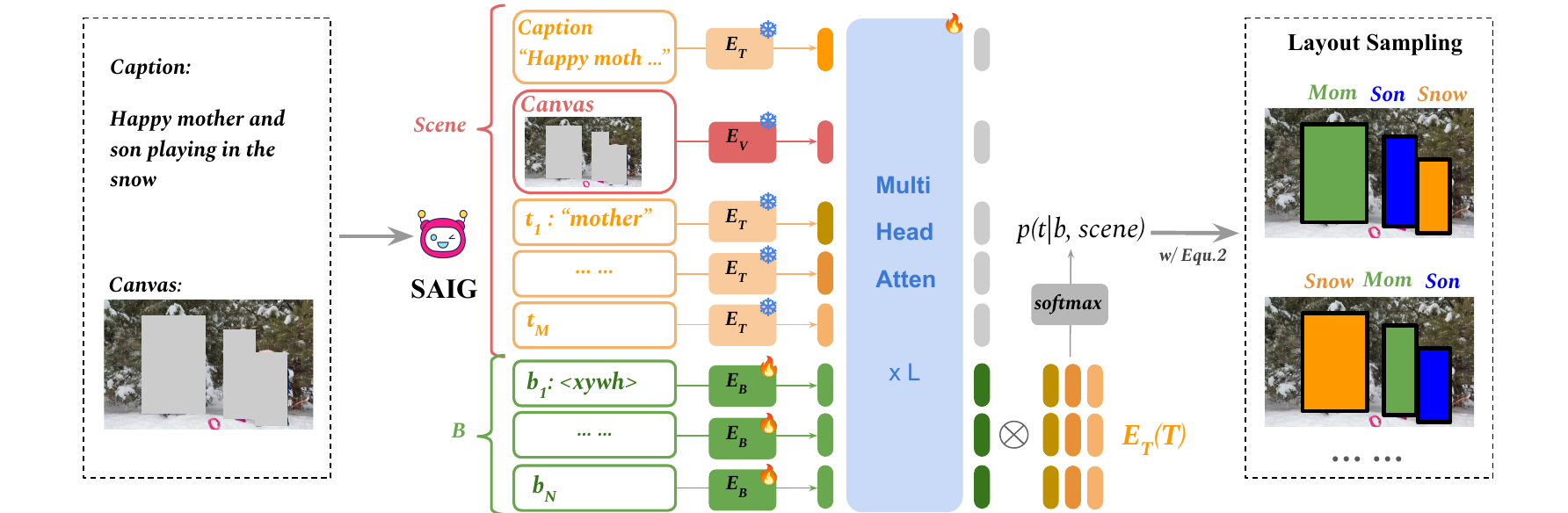}
    \caption{The scene-aware allocation for inpainting is an underexplored but essential task in our generation framework. We propose SAIG to allocate layout with the awareness of the scene.
    }
    \label{fig:saig}
\end{figure*}

In Eq. \ref{eq:4}, $P(t_m | b_n, \text{scene})$ stands for the probability of phrase $t_m$ to be picked for inpainting within $b_n$, while $P(b_n | \text{scene})$ represents $b_n$'s existence in the scene. We parameterize $P(t_m | b_n, \text{scene})$ by a multi-modal multi-layer bidirectional transformer encoder \cite{vaswani2017attention, devlin2018bert}, illustrated in Fig \ref{fig:saig}. 
We engage both visual and textual modalities from the image-caption pair. For the visual modality, we obscure the image within the specified proposal boxes and employ the remaining background as a canvas. 
This canvas prevents the model from gaining knowledge of the original content within the proposal boxes, encouraging it to focus on flexible layout generation.
The caption and canvas are encoded by CLIP textual encoder ($E_T$) and visual encoder ($E_V$), respectively, and projected onto the same visual-semantic space. To facilitate scene understanding, we emphasize some caption phrases that have been extracted in preprocessing, by individually encoding them through $E_T$. The scene is thereby a set of tokens, $\textrm{scene} = \{E_T(\textrm{caption}), E_V(\textrm{canvas}), E_T(t_1), E_T(t_2), ..., E_T(t_m)\}$.

The $b_n$ has a form of <x,y,w,h>. We first encode it through Fourier embedding (FE)\cite{mildenhall2020nerf} and then project it to the same dimension as the other tokens by a trainable multi-layer perception (MLP), formally, $E_B(b_n) = \textrm{MLP}(\textrm{FE}(b_n))$. All encoded modalities are incorporated into the transformer layers with each consisting of a multi-head self-attention block (MHA), an MLP layer, and LayerNorm \cite{ba2016layer}. The output token of the $b_n$ embedding is utilized to conduct dot product with encoded texts, followed by softmax function to calculate the probability that text $t_m$ should be inpainted in $b_n$,

\begin{equation}
\label{eq:7}
     P(t_m | b_n, \textrm{scene}) = \frac{\exp(\textrm{MHA}(E_B(b_n), \textrm{scene}) \cdot E_T(t_m))}{\sum_{j=1}^M \exp(\textrm{MHA}(E_B(b_n), \textrm{scene}) \cdot E_T(t_j))}
\end{equation}

Furthermore, to get $P(b_n | \text{scene})$, we directly leverage the confident score from the class-agnostic detector used in pre-processing to reflect the probability of $b_n$'s existence in the scene. Such wise, we could finally use Eq. \ref{eq:4} to calculate $P(b_n, t_m | \text{scene})$. We use the $P(b_n, t_m | \text{scene})$ to sample diverse and flexible layouts, based on nucleus sampling \cite{holtzman2019curious}.

\subsubsection{Filtering}
The SAIG provides us with allocated layouts that guide image inpainting model to generate region-text pairs. The generated images may contain low-quality regions and it's important to have quality control. We apply two levels of filtering: image level and region level. 
We follow Gen2Det\cite{suri2023gen2det} and run an image aesthetic model\cite{aesthetic} on the generated data. We find that low-scored data is usually low-quality, while very high-scored data is mostly landscape painting and natural scenery, neither are ideal for instance-learning. Additionally, we apply CLIP as a region-level filter on each region-text pair. Appendix \ref{sec:Appendix} provides addition implementation details.


\subsection{Region-to-Text}
\label{sec:r2t}
We further introduce region-to-text generation to augment the textual richness of the region proposals. The image-caption pairs we utilize are mostly sourced from the web, which often results in captions that are erroneous, incomplete, or only partially related to the image subjects.
So a large portion of the original captions only capture merely one or two salient entities instead of mentioning all the semantic details, while some of the captions are simply not directly related to the subject of the image. We aim to fully leverage the potential of these image-caption pair data by generating region-level descriptions via an image captioning model trained in a distinct domain, thus enriching the overall system with semantic details at a granular level. The resulting generated data is both format-compliant and complementary to the text-to-region counterpart.

Given an image $I$, we first obtain regions $\{r_1, r_2, ..., r_N\}$ by cropping the image with enlarged proposal boxes $\mathcal{B}$ which include context for enriched background information. To prevent semantic overlaps and duplicated annotations, all proposal boxes are initially processed by Non-Maximum Suppression. A pre-trained image captioning model \cite{li2023blip} $\mathcal{G}_T$ is applied to generate a set of region-level descriptions $\mathcal{T} = \{t_1, t_2, ... t_N\}$, where prompt is used to guide the model to interpret the image,

\begin{equation}
    \hat{t}_i = \textrm{select}(\mathcal{G_T}(r_i, \textrm{prompt}_1),
    \mathcal{G_T}(r_i, \textrm{prompt}_2),
    \mathcal{G_T}(r_i, \textrm{prompt}_{3})
    )
\end{equation}

In detail, $\hat{t}_i$ are generated by a selecting operation across an ensemble of three text prompt prefixes, for example, "\textit{The image shows} <   >".  As a result, we select the best matching caption for each region proposal according to the highest ranking CLIP similarity score between the region crops and their generated captions. It is predictable that the more prompts we select from, the higher score we will get, but in practice we find three is a good balance between efficiency and effectiveness. 


\subsection{Training OVD with Region-Text Pairs}
\label{sec:lart}
Contrastive learning is recently used in OVD \cite{li2022grounded, kamath2021mdetr, zhou2022detecting, wu2024clim} to force visual features to be similar to their textual features. Here, we first introduce region-text contrastive learning, then we expand it to learn additional object proposals tailored with different localization qualities.

\textbf{Region-text contrastive loss.}
Given an image-caption pair, for $i$-th region $r_i$ we operate ROIAlign\cite{he2017mask} on detector's feature pyramid to extract visual embedding $E_R(r_i)$, and use CLIP pre-trained language model as our text encoder to get the corresponding text embedding $E_T(t_i)$. The pair $(r_i, t_i)$ is recognized as a positive pair. During training, we also maintain a text queue $T^*=\{t^*_l\}_{l \in [L]}$ with a queue length $L$, collected across previous batches. Texts in the queue are assumed dissimilar to $t_i$, and they make the negative pairs with $r_i$, i.e., $\{(r_i, t^*_1)$, $(r_i, t^*_2)$, ..., $(r_i, t^*_L)\}$.  According to \cite{ma2024codet, wu2024clim}, a binary cross-entropy loss is applied,
\begin{equation} 
    \label{eq:rt_loss}
    \mathcal{L}_\textrm{region-text} = -\operatorname{log} \operatorname{\sigma}\left [\tau \cdot \text{cos}(E_R(r_i),E_T(t_i))  \right ] - \sum_{t^*_l \in T^*} \operatorname{log}\left [ 1 - \operatorname{\sigma}\left ( \tau \cdot \text{cos}(E_R(r_i),E_T(t^*_l)) \right ) \right ] \,
\end{equation}


where $\text{cos}$ is cosine similarity, $\tau$ denotes a temperature parameter, and $\sigma$ is sigmoid function.

\textbf{Localization-aware region-text contrastive loss (LART).}
Eq.\ref{eq:rt_loss} only considers aligning $r_i$ and $t_i$, but neglects the importance of precisely localized alignment. As a detector may densely predict many proposals to one single object, it is critical to make the model give the highest confidence rank to the most accurately localized prediction. To involve the awareness of localization quality in contrastive learning, we propose LART. Starting with $(r_i, t_i)$, we first obtain $K$ adjacent regions that overlaps with $r_i$. These adjacent regions can be acquired from the region proposal networks or  dense predictions. We extract their visual embedding $\{E_R(r^*_1), ..., E_R(r^*_K)\}$ and compute their intersection-over-union (IOU) scores $\{s_1, ..., s_K\}$ with $r_i$ as localization quality. 


If a $s_k$ is higher than a predefined threshold $\alpha$, the corresponding $r^*_k$ contains similar information as $r_i$ does, and a positive pair $(r^*_k, t_i)$ is formed. They are trained akin to that of the $(r_i, t_i)$, but their learning loss is down-weighted by $s_k$. This benefits from two perspectives: on one hand, additional positive pairs effectively enlarge the batch size and bring additional supervision; on the other hand, the rescaled loss guarantees the strongest supervision is applied to the origin pair, thus helping the detector confidently predict the optimal localization. If $s_k < \alpha$, the $r^*_k$ contains a relatively small proportion of information from $t_i$. So $r^*_k$ is both negative to $t_i$ and $T^*$. Especially, the negative pair $(r^*_k, t_i)$ distinguishes itself as the $r^*_k$ is derived from $r_i$ rather than from disparate regions, thereby yielding hard-negative examples for more fine-grained learning. Similarly, we obtain:

\begin{equation} 
    \label{equ:}
    \mathcal{L}_\text{adjacent} =  
    -\!\sum_{\substack{k \in [K], \\ s_k\!>\!\alpha}} s_k \operatorname{log} \operatorname{\sigma}\left [\tau\text{cos}(E_R(r^*_k), E_T(t_i))  \right ] -\!  \sum_{\substack{k \in [K], \\ s_k\!<\!\alpha}} \sum_{t_j \in \{t_i, T^*\}} 
    \operatorname{log}\left [ 1\!-\!\operatorname{\sigma}\left ( \tau \text{cos}(E_R(r^*_k),E_T(t_j)) \right ) \right ] 
\end{equation}

The overall objective for LART is thus $\mathcal{L}_\text{LART} = \mathcal{L}_\text{adjacent} + \mathcal{L}_\text{region-text}$.

\textbf{Overall training objective.}
With Faster-RCNN \cite{ren2015faster} and CenterNet2 \cite{zhou2021probabilistic}, we train the detectors parallelly on the detection data $D^{\text{det}}$ and generated data $D^{\text{T2R}}, D^{\text{R2T}}$. Particularly, we treat the image-caption pairs $D^{\text{cap}}$ as a special region-text pair and add them into training. The overall training objective for the detectors is: 
\begin{equation}
\label{eq:1}
\mathcal{L}_{\text{overall}} =
    \begin{cases}
      \mathcal{L}_{\text{rpn}} + \mathcal{L}_{\text{reg}} + \mathcal{L}_{\text{cls}} & \text{if} I\in D^{\text{det}}\\
      \mathcal{L}_{\text{LART}} & \text{if} I \in \{ D^{\text{cap}}, D^{\text{T2R}}, D^{\text{R2T}} \}
   \end{cases} 
\end{equation}

\begin{table}[]    
    \centering
    \caption{\textbf{Comparison with state-of-the-art methods on OV-LVIS benchmark}. CLIP supervision entails the transfer of knowledge from CLIP model. Caption supervision entails the learning of vision-language alignment through image-text pairs. The 'Strict OV' denotes whether the experiments is in strict open-vocabulary setting.}
    \begin{tabular}{lllcc>{\color{gray}}c>{\color{gray}}c>{\color{gray}}c} \toprule
         Method & Supervision & Backbone & Strict OV?& $\text{AP}_{\text{novel}}$ & $\text{AP}_{\text{c}}$ & $\text{AP}_{\text{f}}$ & $\text{AP}_{\text{all}}$  \\
    \midrule
        ViLD\cite{gu2021open}                 & CLIP & RN50-FPN & \ding{51} & 16.6 & 24.6 & 30.3 & 25.5 \\
        RegionCLIP\cite{zhong2022regionclip}  & Caption  & RN50-C4  & \ding{51} & 17.1 & 27.4 & 34.0 & 28.2 \\
        DetPro\cite{du2022learning}           & CLIP  & RN50-FPN & \ding{51} & 19.8 & 25.6 & 28.9 & 25.9 \\
        Detic\cite{zhou2022detecting}         & Caption  & RN50     & \ding{55} & 19.5 & -& - & 30.9 \\
        OV-DETR\cite{zang2022open}            & Caption & RN50-C4  & \ding{55} & 17.4 & 25.0 & 32.5 & 26.6 \\
        PromptDet\cite{feng2022promptdet}     & Caption  & RN50-FPN & \ding{55} & 19.0 & 18.5 & 25.8 & 21.4 \\
        VLDet\cite{lin2022learning}           & Caption & RN50     & \ding{51} & 21.7 & 29.8 & 34.3 & 30.1 \\
        F-VLM\cite{kuo2022f}                  & CLIP  & RN50-FPN & \ding{51} & 18.6 & - & - & 24.2 \\
        BARON\cite{wu2023aligning}            & CLIP  & RN50-FPN & \ding{51} & 22.6 & 27.6 & 29.8 & 27.6 \\
        CoDet\cite{ma2024codet}               & Caption & RN50-FPN & \ding{51} & 23.4 & 30.0 & 34.6 & 30.7 \\
        CLIM\cite{wu2024clim}                 & Caption  & RN50-FPN & \ding{51} & 21.8 & 28.4 & 32.0 & 28.7 \\
        \cellc \textbf{Ours}                         & \cellc \textbf{RTGen}  & \cellc RN50-FPN & \cellc \ding{51}& \cellc \textbf{23.9} & \cellc 28.5 & \cellc 31.8 & \cellc 29.0 \\
        \midrule
        RegionCLIP\cite{zhong2022regionclip} & Caption & R50x4 (87M)  & \ding{51} & 22.0 & 32.1 & 36.9 & 32.3 \\
        Detic\cite{zhou2022detecting}        & Caption & SwinB (88M) & \ding{55} & 23.9 & 40.2 & 42.8 & 38.4 \\
        VLDet\cite{lin2022learning}          & Caption & SwinB (88M) & \ding{51} & 26.3 & 39.4 & 41.9 & 38.1 \\
        F-VLM\cite{kuo2022f}                 & CLIP & R50x4 (87M) & \ding{51} & 26.3 & - & - & 28.5 \\
        CoDet\cite{ma2024codet}              & Caption & SwinB (88M) & \ding{51} & 29.4 & 39.5 & 43.0 & 39.2 \\
        \cellc \textbf{Ours}                        & \cellc \textbf{RTGen}   & \cellc SwinB (88M) & \cellc \ding{51} & \cellc \textbf{30.2} & \cellc 39.9 & \cellc 41.3 & \cellc 38.8 \\
    \bottomrule
    \end{tabular}
    \label{tab:sota_lvis}
    \vspace{-0.1in}
\end{table}

\section{Experiments}
\label{sec:exp}
\subsection{Dataset and Implementation}
\textbf{OV-LVIS} is a widely-used benchmark for OVD on the LVIS dataset\cite{gupta2019lvis}. It contains a total 1203 object categories that are separated into 866 common and frequent objects and 337 rare objects, where the common + frequent are used as base categories and the remaining rare classes as novel objects. The evaluation metric is mask AP of the novel categories. 

\textbf{OV-COCO} is a common benchmark originating from the MSCOCO benchmark \cite{lin2014microsoft, zareian2021open}. Among the 80 categories, 48 are split into the base categories and 17 are reserved as novel categories. The box $\text{AP}_{50}$ on the novel categories is used to evaluate the performance.

\textbf{Conceptual Captions}\cite{sharma2018conceptual} is an image-caption dataset that has $\sim$3.3M web-crawled images paired with the raw text descriptions harvested from Alt-text HTML attribute associated with web images. We conduct our generation process on the available subset ($\sim$2.8M images) once, resulting $\sim$2.0M region-text pairs from the T2R process after filtering, and $\sim$9.6M from the R2T process. 

\textbf{Implementation:}
Unless otherwise specified, for OV-COCO benchmark we conduct experiments using Faster R-CNN with ResNet50, and all models are trained for 45k iterations with AdamW optimizer. For OV-LVIS, we use CenterNet2\cite{zhou2021probabilistic} with ResNet50\cite{He2015DeepRL} and Swin-B\cite{Liu2021SwinTH} trained for 90k iterations. Ablation study is conducted on OV-COCO. Following \cite{zhou2022detecting}, we initialize the models with trained parameters on the base classes to achieve faster convergence. For SAIG, we train it jointly on Flickr30k\cite{Plummer2015Flickr30kEC}, VG Caption\cite{Krishna2016VisualGC} and GQA\cite{Hudson2019GQAAN}. For LART, we simply choose $\alpha=0.5$ as $\text{AP}_{50}$ is a practical metric. 

\subsection{Results on Benchmarks}

Table \ref{tab:sota_lvis} shows our primary results on OV-LVIS. Training with RTGen outperforms other SoTA methods, including those rely on knowledge transfer and mining vision-language alignment from image-caption pairs. This is due to the diversity and semantic richness of generated region-text pairs, and the results is further verified by different backbones. The models supervised by RTGen follow the strict open-vocabulary setting (Section \ref{sec:prelim}) which guarantees the results have no bias to generalize.  

Table \ref{tab:sota_coco} shows our results on the OV-COCO benchmark. The effectiveness of RTGen is consistently demonstrated across multiple OVD benchmarks. As a comparison to effective data augmentation methods, Faster-RCNN model trained with our proposed RTGen supervision outperforms the previous SoTA method CLIM. RTGen also outperforms region-text mining methods such as VLDet and CoDet. 

\begin{table}[t] 
\begin{minipage}[]{0.5\linewidth} 
    \caption{\textbf{Comparison with state-of-the-art on OV-COCO benchmark.}}
    \centering
    \begin{tabular}{lc>{\color{gray}}c>{\color{gray}}c} \toprule
         Method & $\text{AP}^{novel}_{\text{50}}$ & $\text{AP}^{base}_{\text{50}}$ & $\text{AP}^{all}_{\text{50}}$ \\
    \midrule
    OVR-CNN\cite{zareian2021open}        & 22.8 & 46.0 & 39.9 \\
    ViLD\cite{gu2021open}                & 27.6 & 59.5 & 51.3 \\
    RegionCLIP\cite{zhong2022regionclip} & 26.8 & 54.8 & 47.5 \\
    Detic\cite{zhou2022detecting}        & 27.8 & 47.1 & 42.0 \\
    VLDet\cite{lin2022learning}          & 32.0 & 50.6 & 45.8 \\
    CoDet\cite{ma2024codet}              & 30.6 & 52.3 & 46.6 \\
    CLIM]\cite{wu2024clim}               & 32.3 & -    & - \\
    \midrule
    \cellc RTGen (Ours)                         & \cellc \textbf{33.6} & \cellc 51.7 & \cellc 46.9 \\
    \bottomrule
    \end{tabular}
    \label{tab:sota_coco}
\end{minipage}\hfill
    \begin{minipage}{0.45\textwidth}
    \centering
    \caption{\textbf{Ablation study of RTGen on OV-COCO.} The models are trained with $\mathcal{L}_{region-text}$ if w/o LART.} 
    \setlength{\tabcolsep}{4pt}
    \resizebox{\columnwidth}{!}{
    \begin{tabular}{ccccc} \toprule
         CC3M & T2R & R2T & LART & $\text{AP}^{novel}_{\text{50}}$ \\
         \midrule
                    &           &               &               &  \textcolor{gray}{2.9}  \\ 
         \ding{51}  &           &               &               &       25.7              \\
         \ding{51}  & \ding{51} &               &               &       29.2(+3.5)        \\
         \ding{51}  &           &   \ding{51}   &               &       28.1(+2.4)        \\
         \ding{51}  & \ding{51} &   \ding{51}   &               &       31.4(+4.7)        \\ 
                    & \ding{51} &  \ding{51}    &  \ding{51}    &       31.6(+4.9)        \\
         \ding{51}  & \ding{51} &   \ding{51}   &  \ding{51}    &       \textbf{33.6(+6.9)}    \\ 
    \bottomrule
    \end{tabular} 
    \label{tab:ablation_data_impact}
    }
\end{minipage}
\end{table}


\begin{table}[b] 
\begin{minipage}[]{0.5\linewidth} 
    \caption{\textbf{Generating vs. Pseudo-Labeling}}
    \setlength{\tabcolsep}{4pt}
    \centering
    \begin{tabular}{ccccc} \toprule
         CC3M       & LART      & Pseudo       & T2R+R2T       & $\text{AP}^{novel}_{\text{50}}$ \\
         \midrule
         \ding{51}  &           &               &               &       25.7      \\ 
         \ding{51}  & \ding{51} &   \ding{51}   &               &       31.2      \\    
         \ding{51}  & \ding{51} &               &   \ding{51}   &       33.6      \\ 
    \bottomrule
    \end{tabular} 
    \label{tab:pseudol}
\end{minipage}\hfill
    \begin{minipage}{0.41\textwidth}
    \centering
    \caption{\textbf{Queue Length of LART}} 
    \resizebox{\columnwidth}{!}{
    \begin{tabular}{ccc}
    \toprule
         Queue Length  & $\text{AP}^{novel}_{\text{50}}$ & $\text{AP}^{novel}_{\text{75}}$ \\
    \midrule
         $L=512$ & 32.0 & 18.9\\
         $L=125$ & 32.5 & 19.6\\
         $L=256$ & 33.6 & 20.1\\
    \bottomrule
    \end{tabular}
    \label{tab:lartweight}
    }
\end{minipage}
\end{table}

\subsection{Ablation Studies}

\textbf{Impact of the generated data}\quad
To understand the effectiveness of different sources of supervision, we conduct experiments on the OV-COCO benchmark with various combinations of data, as shown in Table \ref{tab:ablation_data_impact}. Our baseline is the Detic\cite{zhou2022detecting} trained with image-caption pairs (CC3M) by treating entire image as region. The models further trained with region-text supervision show continuous improvements (+3.5 from T2R and +2.4 from R2T). Both R2T and T2R can work jointly and bring the improvement to a new high (+4.7). Additionally, Figure \ref{fig:fig1}(c) illustrates a consistent improvement of generated data with an increasing percentage of data involved, suggesting its scalability. 

\textbf{Impact of LART}\quad
LART involves the awareness of localization in contrastive learning. As shown in Table \ref{tab:ablation_data_impact}, the integration of LART yields a significant contribution of an additional +2.2AP to the already high results (31.6).  Findings in Table \ref{tab:lartweight} indicate that the queue length influences the training outcomes: a longer queue length exacerbates the learning challenges, disrupting the equilibrium between localization learning and open-vocabulary recognition. Conversely, a short queue length is also not ideal for contrastive learning.

\subsection{Generating Region-Text Pairs vs. Pseudo Labeling}
We compare generating region-text pairs with the pseudo-labeling.  The pseudo labels are predicted by a well-trained Grounding DINO\cite{liu2023grounding} on the same CC3M dataset, and the resulting region-text pairs are then used following the same detection training recipe. In Table \ref{tab:pseudol}, pseudo-labeling is inferior to generating. This can be attributed to two primary factors: 1. generating derives from two distinct processes that involve separate models from different domains, thereby enhancing the diversity and semantic richness of the data.  2. the generation framework remains robust even when the alignment between image-caption pairs is suboptimal, in contrast to pseudo-labeling, which is significantly compromised under such circumstances.

\begin{figure*}[t]
    \centering
    \includegraphics[width=1.0\columnwidth]{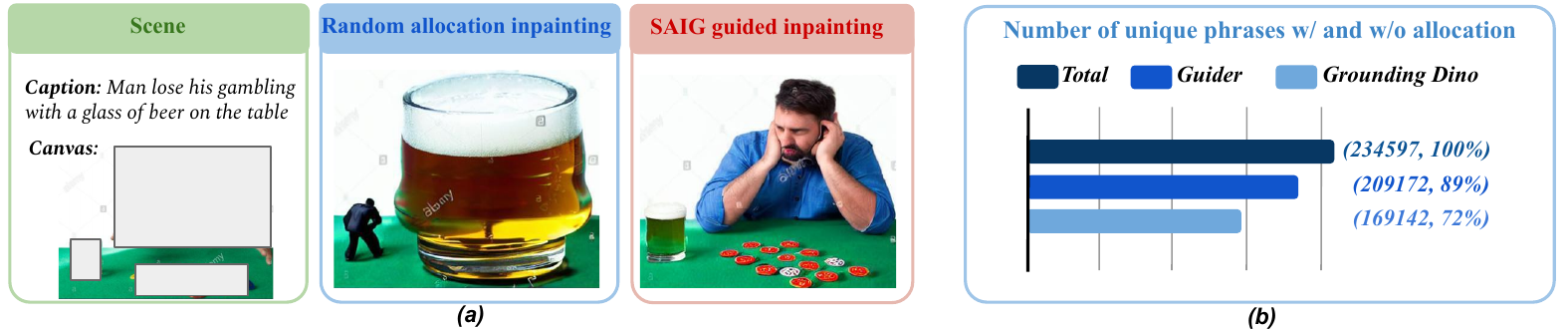}
    \caption{(a) SAIG allocates the phrases "Man" and "a glass of beer" to more suitable boxes according to the scene. (b) SAIG preserves more semantics information during allocation than grounding.}
    \label{fig:randg}
\end{figure*}

\subsection{Comparing Alternative Allocation Methods for Inpainting}

\begin{wrapfigure}{r}{0.32\textwidth}
    \vspace{-0.4in}
    \centering
        \includegraphics[width=0.3\textwidth]{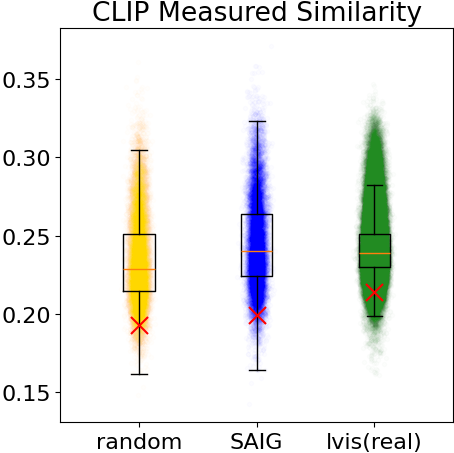}
        \vspace{-0.1in}
        \caption{Alignment quality of pairs. $\textcolor{red}{\times}$ indicates the low $3\%$.}
        \label{fig:ablation_clip_sim}
\end{wrapfigure}

We analyze SAIG by comparing it with two alternative designs. 

\textbf{Design (a): Randomly allocating} $\mathcal{B}$ and $\mathcal{T}$ is straightforward but overlooks the intrinsic geometric property and graphical relation, leading to hilarious outcomes (see Figure \ref{fig:randg}(a)). We quantify the quality by measuring the CLIP similarity of generated 10K region-text pairs in Figure \ref{fig:ablation_clip_sim}, where the results on the LVIS val set are also presented as an upper bound . Clearly, SAIG leads to better inpainting quality due to the reasonable layout guidance.

\textbf{Design (b): Grounding} applies a pre-trained grounding model to pair the regions with the phrases from the caption. However, our practical finding reveals that grounding fails to capture a significant proportion of phrases during the allocation, thereby reducing the semantic diversity. As depicted in Figure \ref{fig:randg}(b), when comparing with Grounding DINO that pseudo-labels region-text allocations, our method surpasses in capturing more noun phrases. This comparison underscores the limitation of grounding, which frequently overlook phrases due to low confidence scores and phrases absent in corresponding images.


\section{Conclusions and Limitations}
In this work, we propose to generate region-text pairs for training open-vocabulary object detection. Our approach innovates text-to-region and region-to-text processes, along with the introduction of the Scene-Aware Inpainting Guider and the Localization-Aware Region-Text Contrastive Loss. Experiments on OV-COCO and OV-LVIS benchmarks show that detectors trained with generated data outperform current OVD approaches, demonstrating its effectiveness. Moreover, our method is orthogonal to previous methods that focus on achieving region-text alignments, making generation methods a promising approach to reach good performance. We recognize that training downstream computer vision tasks with generated data is still not well explored and we encourage further investigations in this promising direction.

{
\small
\bibliographystyle{plain}
\bibliography{egbib}
}

\appendix

\section{Appendix} 
\label{sec:Appendix}
\subsection{Implementation Detail of the Scene-Aware Inpainting Guider}
The SAIG is constructed with 32 layers of multi-head self-attention.  We use CLIP-Vit-L/14 as a feature extractor. The box encoder contains three full-connected layers with SiLU activation function in between. The cross-entropy loss is applied for training. AdamW with learning rate=1e-4 is chosen as optimizer. We train the guider with 8xA100 GPUs for 12 epochs until it converges. 

\subsection{Implementation Detail of the Proposal Box Generator}
We look for regions of interest in the images before applying the generation models. Specifically, we use an off-the-shelf class-agnostic object proposal generator MViTs\cite{Maaz2022mvit} to predict object proposals with the text prompts "\textit{all objects}" and "\textit{all entities}". Regions with a confidence score above 0.3 are kept and ensembled. To avoid repetitive region proposals, all regions are first filtered by the NMS process with a 0.1 IoU threshold. 

\subsection{Implementation Detail of Phrases Parser}
We use a large language model Mistral\cite{jiang2023mistral} and NLTK \cite{loper2002nltk} word-tree to extract phrases that are suitable for inpainting from captions.  Specifically, we find directly using a prompt like "please list tangible objects in the sentence" often produces sub-optimal results and gives incorrect phrases such as "beauty", "university", "sunday", and "nightmare". Therefore we choose an intruct-finetuned variant Mistral-8x7B-Instruct-v0.2, where we prompt several examples and use Prompt+Instruct \cite{chen2024autoprm} for in-context learning. The selected examples and prompt template are shown in Table \ref{tab:LLM}. 

Afterwards, we use a word-tree to filter the extracted phrases by the hierarchy with allowance and forbidden categories, summarized in Table \ref{tab:nltk}.  If a phrase's hypernym appears or disappears in both categories, it will be dropped. 

\begin{table}[htbp]
\centering
\caption{Prompting LLM for Phrase Extraction}
\label{tab:LLM}
\begin{tabular}{l}
\toprule
\textbf{Prompt:}  \\ \midrule
Export the real-world objects with a physical body in the sentence, return None if not found. \\ \midrule

\textbf{Instruct:}   \\ \midrule
User: burger: pound of fries and some sauces, man talking on his smart phone on the beach in \\ cloudy dark weather. Assistant: burger, fries, some sauces, man, smart phone.   \\ \midrule
User: medical team working together at night, taking care of patients carefully on a hospital \\ ward.  Assistant: mediacal team, patients.  \\  \midrule
User: night display of sculptures during olympic games. Assistant: sculptures. \\  \midrule
User: where is the sea in space?. Assistant: None       \\
\bottomrule
\end{tabular}
\end{table}

\begin{table}[htbp]
\centering
\caption{NLTK Hierarchy Hypernym Filter}
\label{tab:nltk}
\begin{tabular}{>{\raggedright\arraybackslash}p{6cm}|>{\raggedright\arraybackslash}p{6cm}}
\toprule
\textbf{Allowance} & \textbf{Forbidden}  \\
\midrule
'physical entity', 'food', 'person', 'living thing', 'social group', 'biological group'  
& 
'measure', 'atmosphere', 'time', 'activity', 'phenomenon', 'event', 'meeting',  'organization', 'location', 'land', 'facility' \\ 
\bottomrule
\end{tabular}
\end{table}

\subsection{Filtering Generated Data}
As explained, the generated images may contain low-quality regions and we want to filter out these low quality generation before the training of the detectors. We image-level filtering and region-level filtering. We apply an aesthetic filter \cite{aesthetic} and select the 95 percentile interval threshold $t_1$ and $t_2$ for all images. The images with aesthetic scores outside of the range  $(t_1, t_2)$ are filtered out. In our final implementation, we select $t_1 = 3.0$ and $t_2=6.0$. Note that we also remove images with high aesthetic scores because most of them contain natural scenery, which might not be ideal for region-text alignment learning. Subsequently, we apply an adaptive region-level filter to remove inpainted regions with poor quality and we use a pretrained CLIP model \cite{radford2021learning} as a filter. Concretely, for a generated region-text pair, we calculate the cosine similarity scores between the region and all the text phrases. A region annotation will be filtered out if the similarity score between the region and the correspondent text phrases is less than the top 5\% of all the text phrases. Empirically, we find the dynamic threshold works better than a fixed threshold as it preserves text phrases that might have multiple synonyms.

\subsection{OVD Implementation Details}
We highlight the hyperparameters and training configuration used for the OV-COCO and OV-LVIS benchmark experiments. Following Detic \cite{zhou2022detecting} and CoDet \cite{ma2024codet}, we use a dataset ratio of 1:4 between detection/region-text dataset for fair comparison. We consider the original CC3M \cite{sharma2018conceptual} image-caption data as a generalized case for the region-text pair data and concatenate it with R2T data. Table \ref{tab:configuration} lists all details during training. We use the well-known academic repository MMDetection \cite{mmdetection} to build our training pipeline. All experiments are conducted on a single server node with 8 Nvidia A100 GPUS. 

\begin{table}[htbp]
\centering
\caption{Hyperparameters for training OVD on OV-LVIS and OV-COCO.}
\label{tab:configuration}
\begin{tabular}{@{}lcc@{}}
\toprule
Benchmark                        & OV-LVIS                & OV-COCO                      \\ \midrule
Detection Framework              & CenterNet2             & Faster RCNN                   \\
Optimizer                        & AdamW                  & SGD                          \\
Gradient clipping                & True                   & True                         \\
Learning rate (LR)               & 2e-4                   & 2e-2                         \\
Total iterations                 & 90k                    & 45k                          \\
Warmup iterations                & 10k                    & 2k                           \\
Decay schedule                   & CosineAnnealing        & StepDecay[30k, 40k]          \\
Step decay factor                & --                     & 0.1$\times$                  \\
Data augmentation                & LSJ, HorizontalFlip    & HorizontalFlip               \\
Batch size per GPU (detection)   & 8                      & 2                            \\
Batch size per GPU (R2T)         & 16                     & 4                            \\
Batch size per GPU (T2R)         & 16                     & 4                            \\
Resolution (detection)           & 896 $\times$ 896       & 1333 $\times$ 800            \\
Resolution (caption)             & 448 $\times$ 448       & 667 $\times$ 400             \\
Federated loss                   & True                   & False                        \\
Repeat factor sampling           & True                   & False            \\
$\mathcal{L}_\text{LART}$ weight & 1.0                    & 1.0                          \\ \bottomrule
\end{tabular}
\end{table}

\subsection{Visualizations}
We visualize our generated region-text pair as well as the detection results. For region-text pair, each region is highlighted by a blue bounding box and green text. On OV-COCO benchmark, we show detection results of both the base classes and novel classes. On OV-LVIS, we can see that the detector is able to generalize to many rare categories such as \textit{"clementine", "harmonium", "joystick", "boom\_microphone"}, etc.

\begin{figure}[ht]
    \includegraphics[width=.33\textwidth]{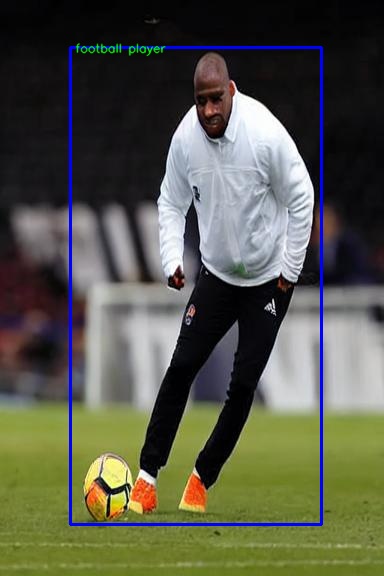}\hfill
    \includegraphics[width=.37\textwidth]{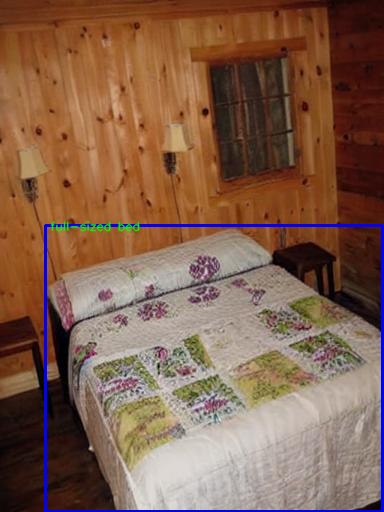}\hfill
    \includegraphics[width=.28\textwidth]{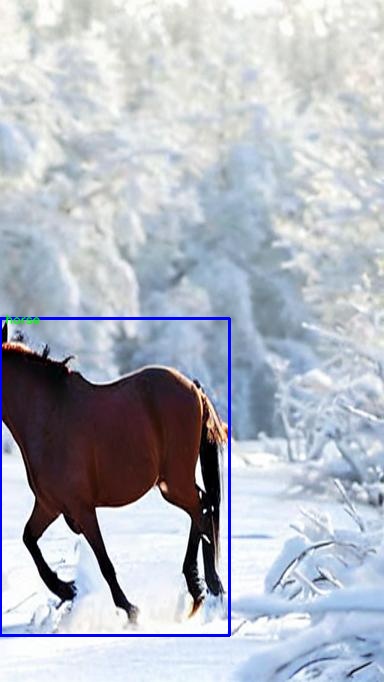}
    \\[\smallskipamount]
    \includegraphics[width=.44\textwidth]{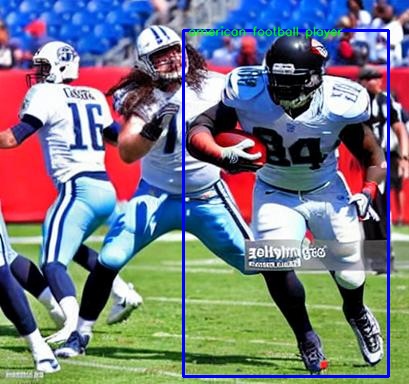}\hfill
    \includegraphics[width=.54\textwidth]{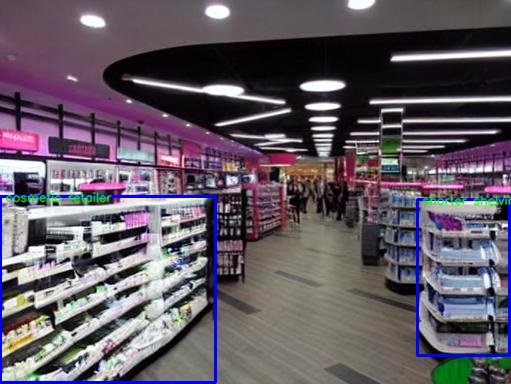}
    \\[\smallskipamount]
    \includegraphics[width=.42\textwidth]{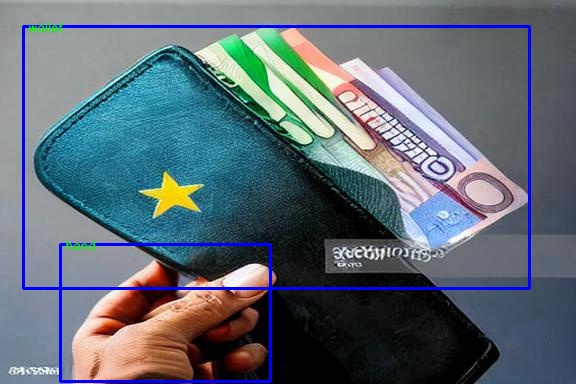}\hfill
    \includegraphics[width=.19\textwidth]{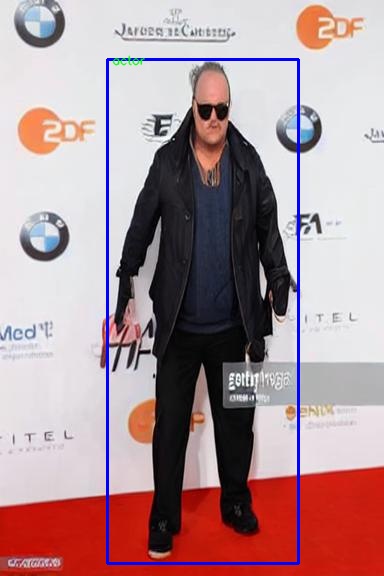}\hfill
    \includegraphics[width=.38\textwidth]{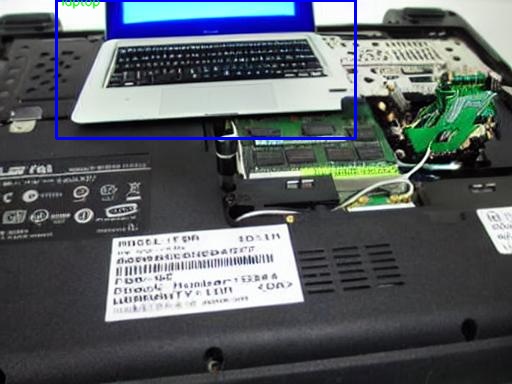}
    \\[\smallskipamount]
    \includegraphics[width=.48\textwidth]{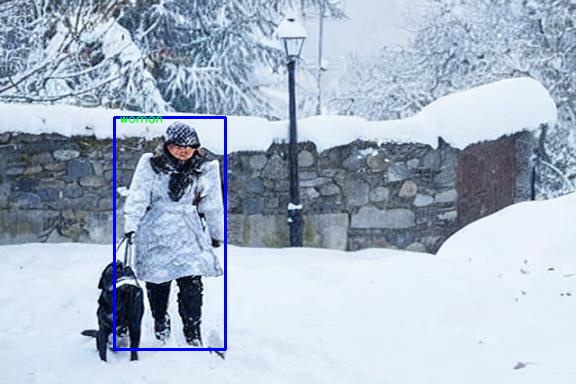}\hfill
    \includegraphics[width=.5\textwidth]{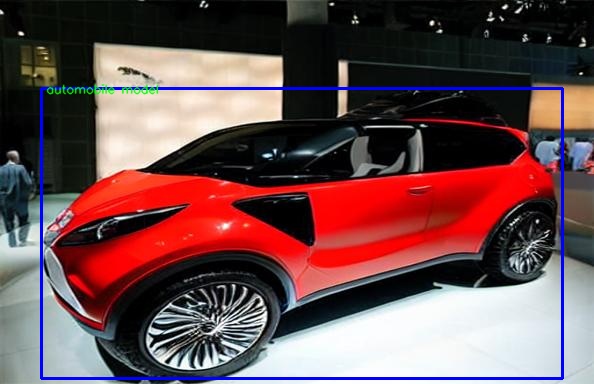}
    \caption{\textbf{Examples of T2R generation}, zoom in better visualization.}\label{fig:t2rvis}
\end{figure}

\begin{figure}[ht]
    \includegraphics[width=.28\textwidth]{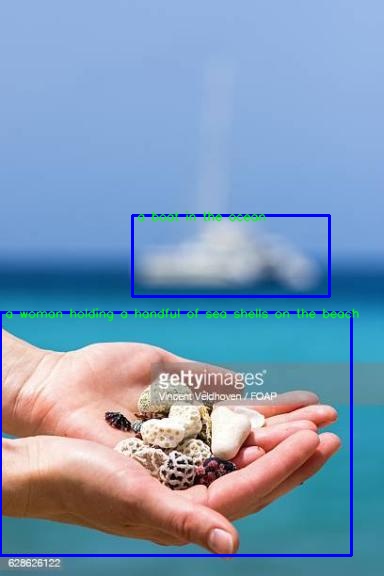}\hfill
    \includegraphics[width=.40\textwidth]{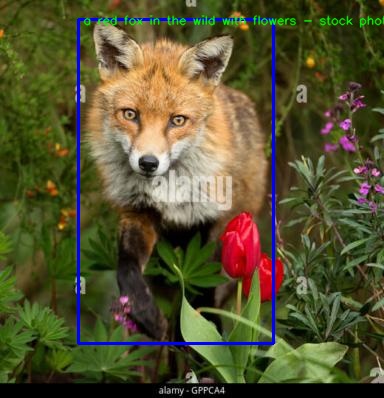}\hfill
    \includegraphics[width=.28\textwidth]{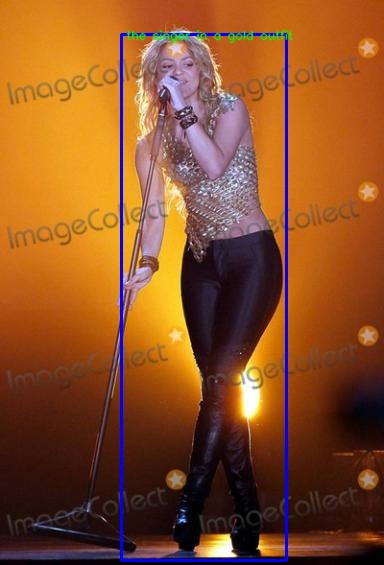}
    \\[\smallskipamount]
    \includegraphics[width=.45\textwidth]{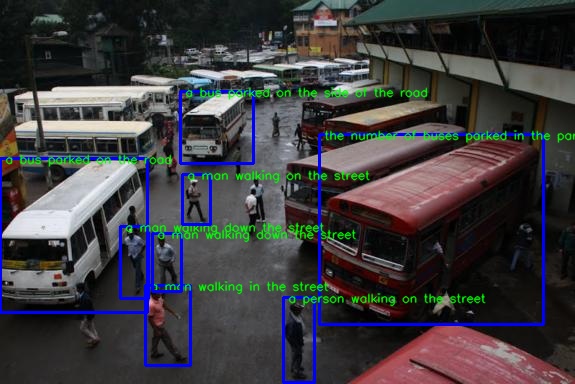}\hfill
    \includegraphics[width=.53\textwidth]{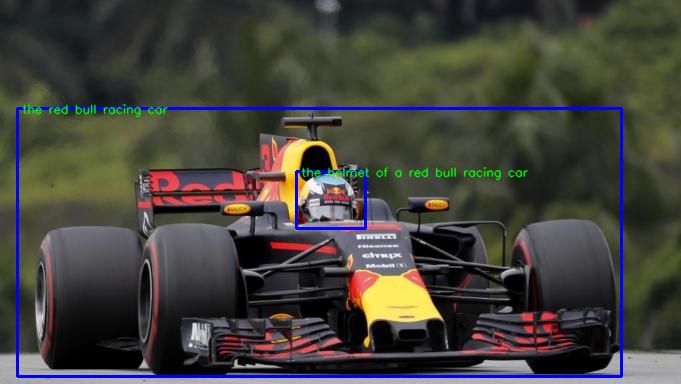}
    \\[\smallskipamount]
    \includegraphics[width=.33\textwidth]{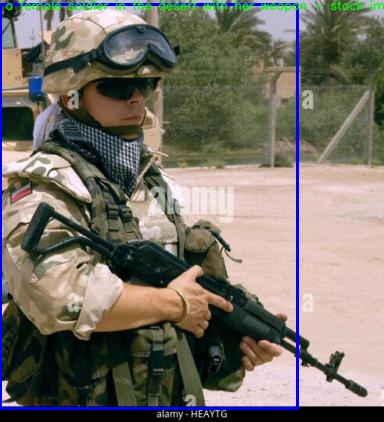}\hfill
    \includegraphics[width=.65\textwidth]{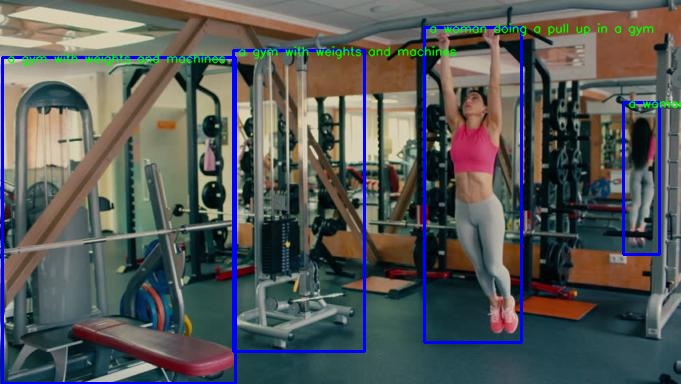}
    \\[\smallskipamount]
    \includegraphics[width=.332\textwidth]{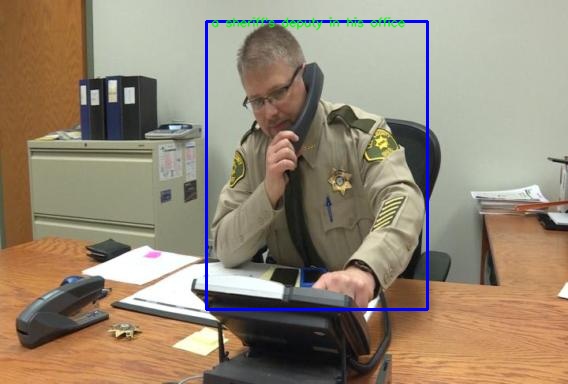}
    \includegraphics[width=.325\textwidth]{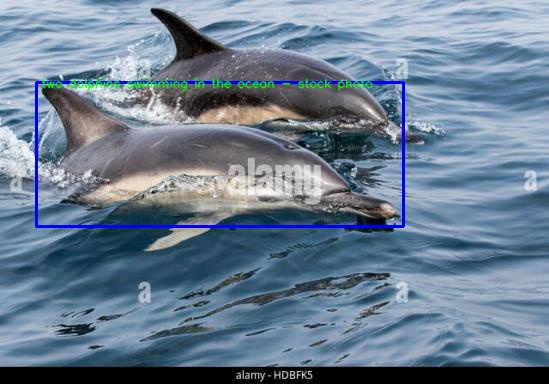}
    \includegraphics[width=.325\textwidth]{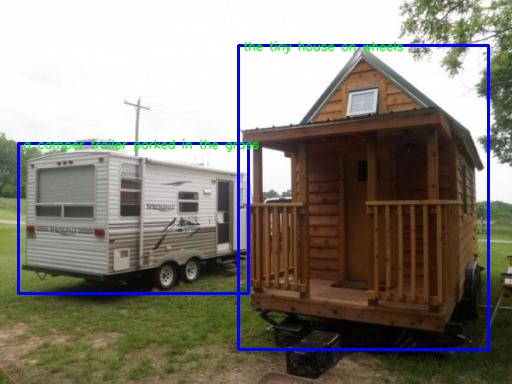}
    \caption{\textbf{Examples of R2T generation}, zoom in better visualization.}\label{fig:r2tvis}
\end{figure}

\begin{figure}[ht]
    \includegraphics[width=.45\textwidth]{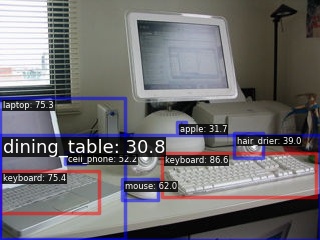}\hfill
    \includegraphics[width=.25\textwidth]{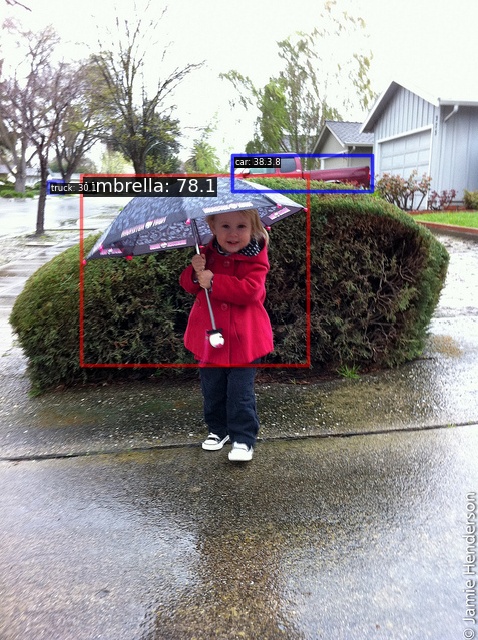}\hfill
    \includegraphics[width=.225\textwidth]{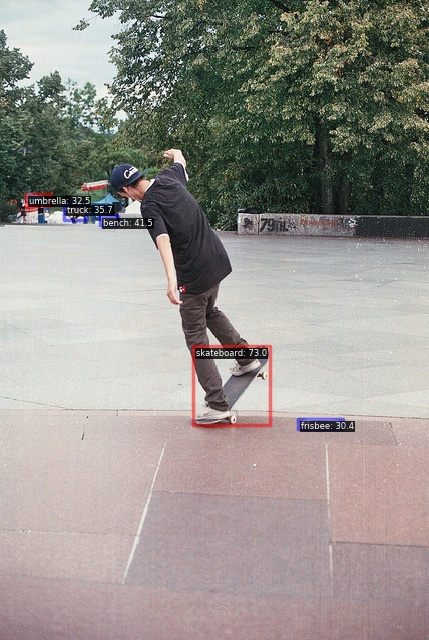}
    \\[\smallskipamount]
    \includegraphics[width=.47\textwidth]{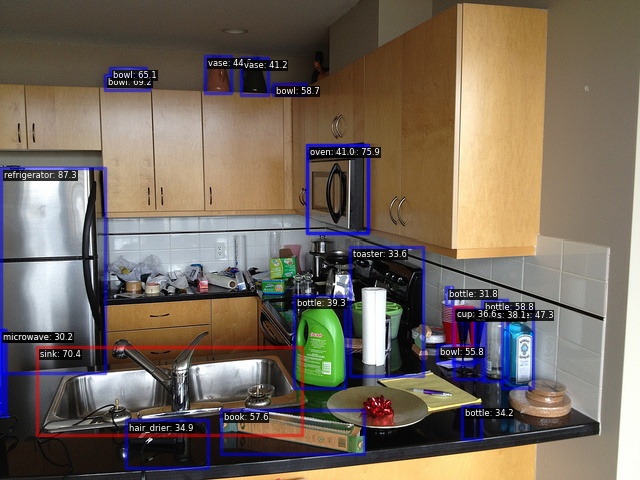}\hfill
    \includegraphics[width=.52\textwidth]{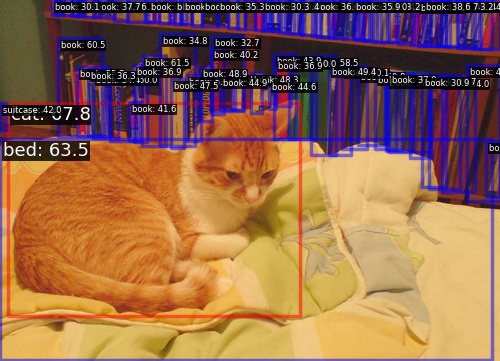}
    \\[\smallskipamount]
    \includegraphics[width=.32\textwidth]{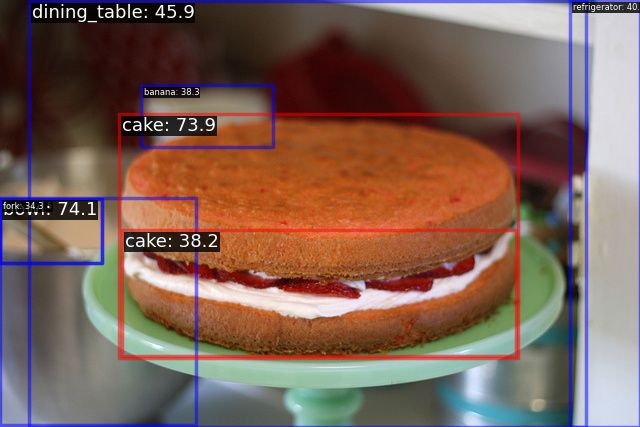}\hfill
    \includegraphics[width=.16\textwidth]{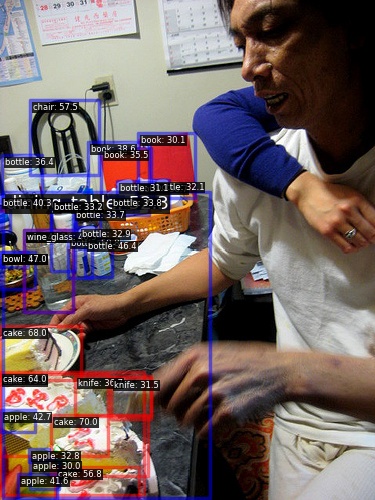}\hfill
    \includegraphics[width=.32\textwidth]{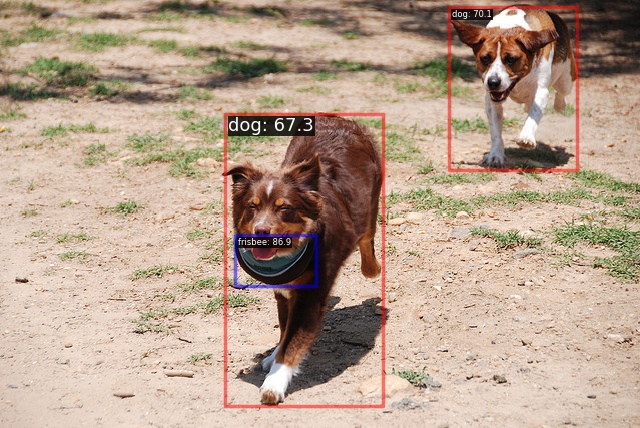}\hfill
    \includegraphics[width=.16\textwidth]{coco_000000311180.jpg}
    \\[\smallskipamount]
    \includegraphics[width=.21\textwidth]{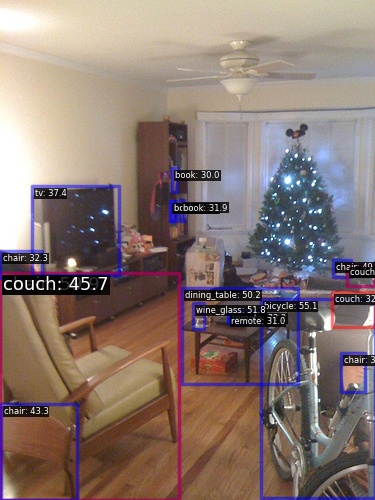}\hfill
    \includegraphics[width=.35\textwidth]{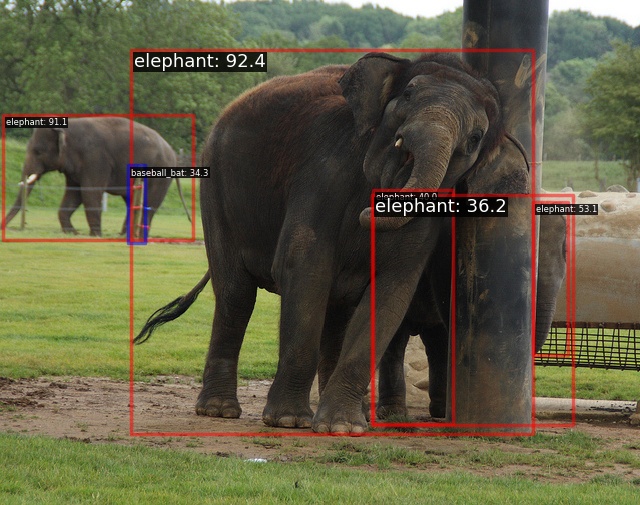}\hfill
    \includegraphics[width=.41\textwidth]{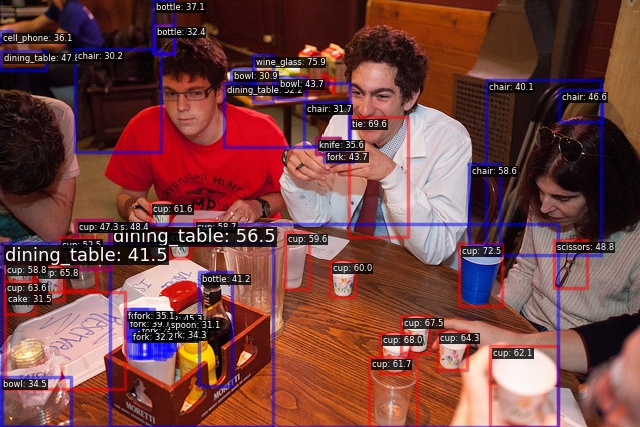}\hfill
    \caption{\textbf{Visualization of detection results on OV-COCO}, base categories are shown in blue boxes while red boxes are for novel categories.}
    \label{fig:coco_visualization}
\end{figure}

\begin{figure}[ht]
    \includegraphics[width=.275\textwidth]{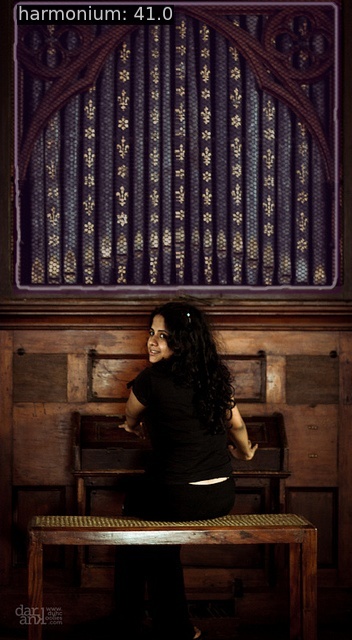}
    \includegraphics[width=.335\textwidth]{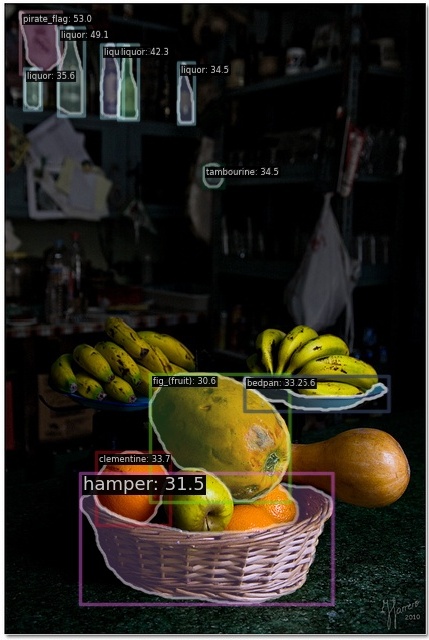}
    \includegraphics[width=.37\textwidth]{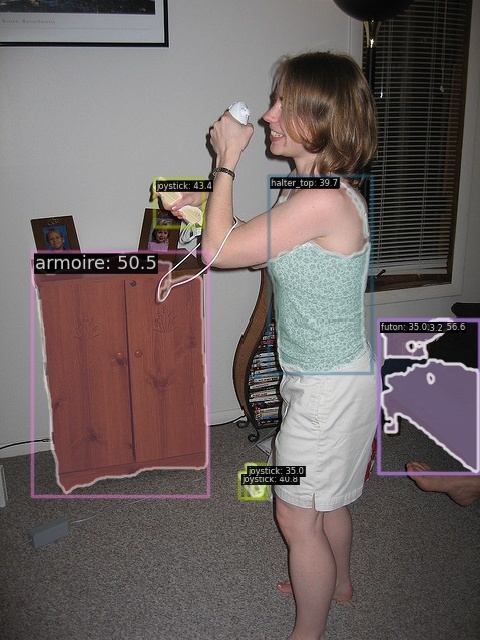}
    \\[\smallskipamount]
    \includegraphics[width=.26\textwidth]{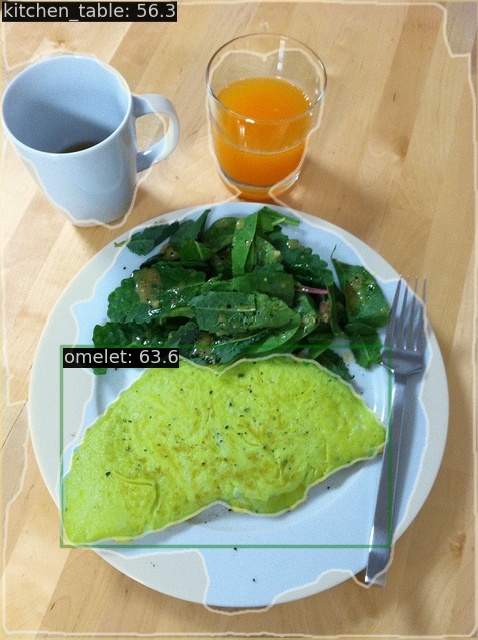}
    \includegraphics[width=.46\textwidth]{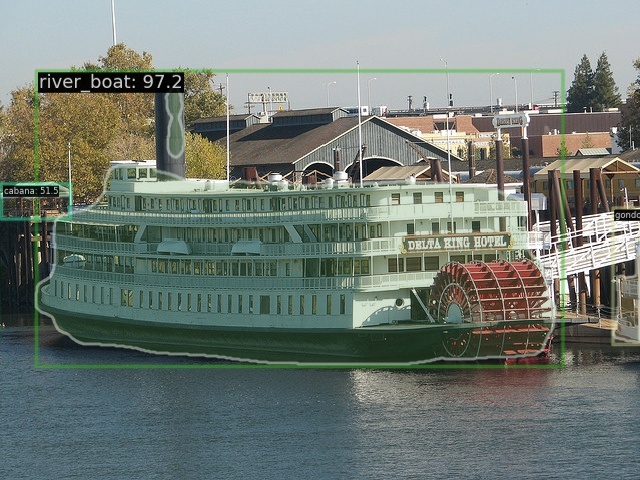}
    \includegraphics[width=.26\textwidth]{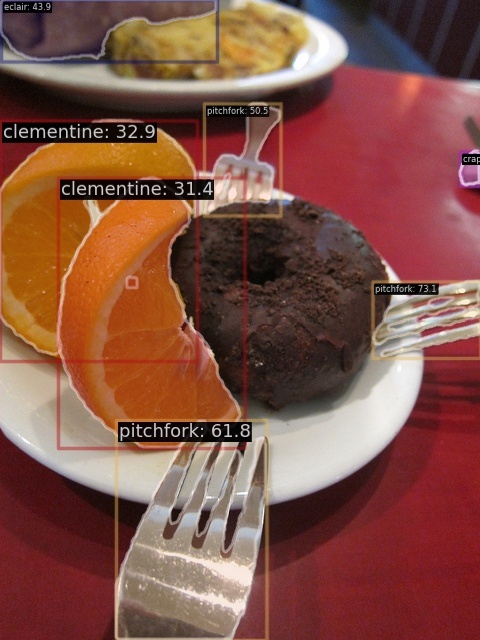}
    \\[\smallskipamount]
    \includegraphics[width=.34\textwidth]{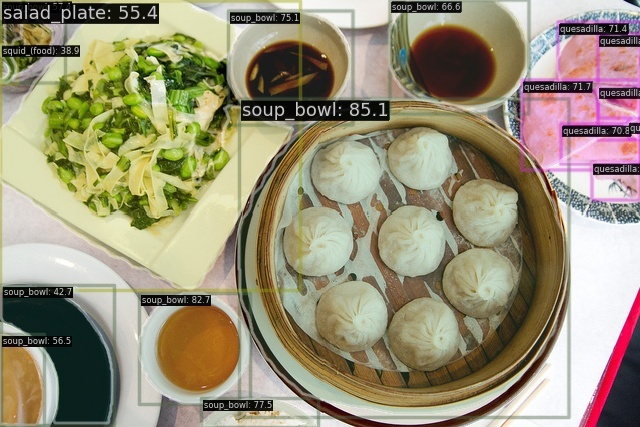}
    \includegraphics[width=.35\textwidth]{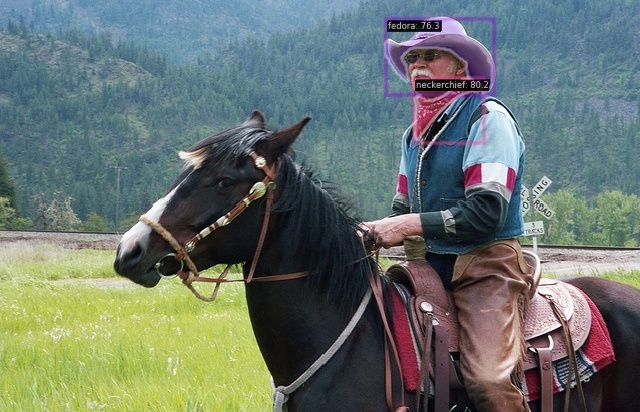}
    \includegraphics[width=.3\textwidth]{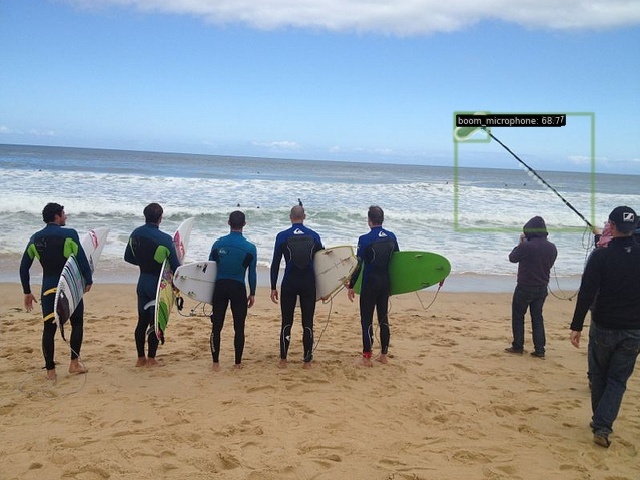}
    \\[\smallskipamount]
    \includegraphics[width=.46\textwidth]{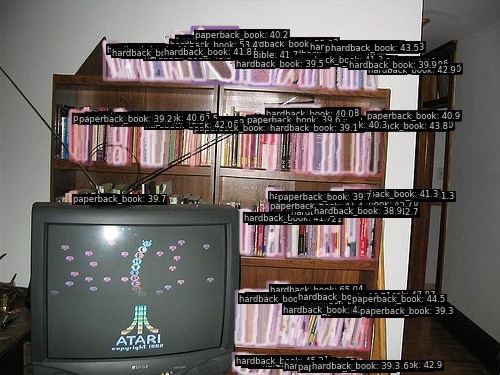}\hfill
    \includegraphics[width=.52\textwidth]{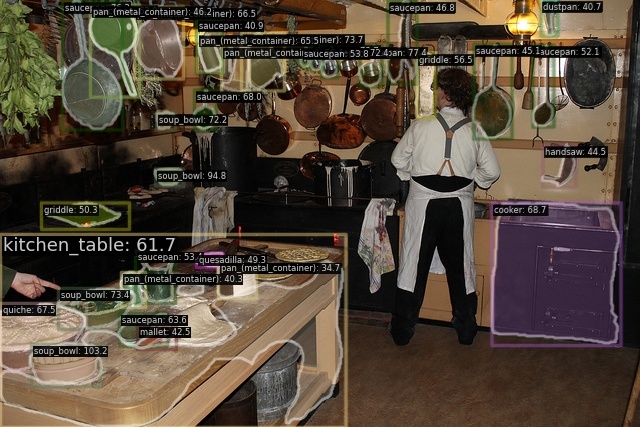}
    \caption{\textbf{Visualization of detection results on OV-LVIS}, to avoid clutteredness, only the novel categories are shown.}\label{fig:lvis_visualization}
\end{figure}

\clearpage

\subsection{Notations and Acronyms}
\begin{table}[htbp]
\centering
\caption{A list of all notations}
\label{tab:notations}
\begin{tabular}{ll}
\toprule
Notation                        & Description                       \\ \midrule
$C^{\text{base}}$               & base categories                   \\
$C^{\text{novel}}$              & novel categories                   \\
$C^{\text{open}}$               & open vocabulary categories         \\
T2R                             & text to region generation          \\
T2T                             & region to text generation           \\
$D^{\text{det}}$                & detection dataset              \\
$D^{\text{cap}}$                & image-caption pair dataset              \\
$D^{\text{T2R}}$                & text to region generation dataset              \\
$D^{\text{R2T}}$                & region to text generation dataset              \\
$r_i$                           & $i$-th image region                \\
$\hat{r}_i$                     & $i$-th inpainted image region            \\
$r^*_k$                         & $k$-th adjacent region proposal           \\
$t_i$                           & $i$-th text phrase                       \\
$\hat{t}_i$                     & $i$-th generated text phrase                  \\
$t^*_l$                         & $l$-th text phrase in text queue  $T^*$          \\
$\mathcal{G_I}$                 & inpainting model                      \\
$\mathcal{G_T}$                 & captioning model                      \\
$b_i$                           & $i$-th proposal box                        \\
$I$                             & image                              \\
$\mathcal{B}$                   & set of proposal boxes               \\
$\mathcal{T}$                   & set of extracted text phrases         \\
$p_{nm}$                        & joint probability to allocate a pair $(b_n, t_m) $ \\
$E_T$                           & CLIP textual encoder                  \\
$E_V$                           & CLIP visual encoder                  \\
$E_R$                           & region encoder from detectors         \\
$T^*$                           & text queue for contrastive learning        \\
$L$                           & text queue length        \\
LART                            & Localization-aware region-text contrastive loss \\
$\sigma$                        & sigmoid function                      \\
$\tau$                          & temperature parameter used in contrastive loss \\
$s_k$                           & localization quality of the $k$-th adjacent region   \\
$\alpha$                        & localization quality threshold \\
\bottomrule
\end{tabular}
\end{table}
\end{document}